\pdfoutput=1

\documentclass[11pt]{article}

\usepackage[final]{acl}

\usepackage{times}
\usepackage{latexsym}

\usepackage[T1]{fontenc}

\usepackage[utf8]{inputenc}

\usepackage{microtype}

\usepackage{inconsolata}

\usepackage{graphicx}
\usepackage{algorithm}
\usepackage{algorithmic}
\usepackage{amsmath, amssymb, amsthm}
\usepackage{tabularx}
\usepackage{color}
\usepackage{fancyvrb}   
\usepackage[breakable]{tcolorbox}
\usepackage{listings}
\usepackage{multirow}
\usepackage{authblk}

\title{Cultivating Game Sense for Yourself: Making VLMs Gaming Experts}

\author{Wenxuan Lu\textsuperscript{1,2},
 Jiangyang He\textsuperscript{3},
 Zhanqiu Zhang\textsuperscript{4}$^{\dagger}$,
 Yiwen Guo\textsuperscript{5}$^{\dagger}$,
Tianning Zang\textsuperscript{1,2}$^{\dagger}$
 \\
  Institute of Information Engineering, Chinese Academy of Sciences\textsuperscript{1}\\
  School of Cyber Security, University of Chinese Academy of Sciences\textsuperscript{2}\\
  School of Computer Science and Artificial Intelligence, Wuhan University of Technology\textsuperscript{3}\\
  LIGHTSPEED\textsuperscript{4}\\
  Independent Researcher\textsuperscript{5}\\
  luwenxuan@iie.ac.cn, he\_jiang\_yang@whut.edu.cn,  zqzhang27@gmail.com, \\ zangtianning@iie.ac.cn, guoyiwen89@gmail.com}

\begin{document}
\maketitle
\begin{abstract}

Developing agents capable of fluid gameplay in first/third-person games without API access remains a critical challenge in Artificial General Intelligence (AGI). Recent efforts leverage Vision Language Models (VLMs) as direct controllers, frequently pausing the game to analyze screens and plan action through language reasoning. However, this inefficient paradigm fundamentally restricts agents to basic and non-fluent interactions: relying on isolated VLM reasoning for each action makes it impossible to handle tasks requiring high reactivity (e.g., FPS shooting) or dynamic adaptability (e.g., ACT combat). To handle this, we propose a paradigm shift in gameplay agent design: instead of directly controlling gameplay, VLM develops specialized execution modules tailored for tasks like shooting and combat. These modules handle real-time game interactions, elevating VLM to a high-level developer. Building upon this paradigm, we introduce GameSense, a gameplay agent framework where VLM develops task-specific game sense modules by observing task execution and leveraging vision tools and neural network training pipelines. These modules encapsulate action-feedback logic, ranging from direct action rules to neural network-based decisions. Experiments demonstrate that our framework is the first to achieve fluent gameplay in diverse genres, including ACT, FPS, and Flappy Bird, setting a new benchmark for game-playing agents.



\end{abstract}

\section{Introduction}
\footnote{${\dagger}$ Corresponding authors}
Developing agents that fluidly play first/third-person games without API access remains a critical challenge in AGI, where complexity mirrors real-world embodied tasks ~\cite{lu2024chameleon, wang2024survey}. Agents must navigate diverse tasks, ranging from combat encounters to environmental navigation, while executing precise real-time actions ~\cite{hu2024survey}. Traditional reinforcement learning (RL) approaches struggle to handle such a broad spectrum of demands due to their limited task generalization ~\cite{de2022automated,justesen2019deep}. Recently, the emergence of Vision Language Models (VLMs) has opened new possibilities in this domain. With their strengths in visual understanding and decision-making, VLMs interact with games purely through visual understanding of game screens. This ability offers a promising direction for developing non-API-dependent gameplay agents ~\cite{tan2024cradle, liu2024rl, wang2023describe}.

\begin{figure}[t]
    \centering
    \includegraphics[width=\linewidth]{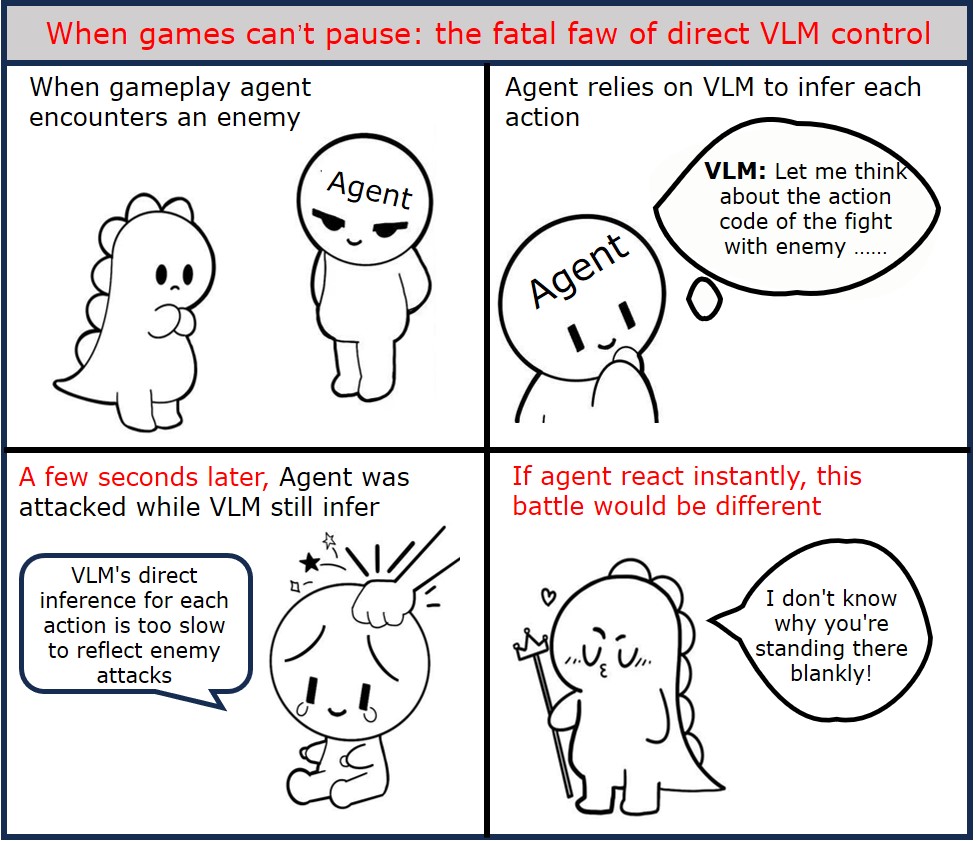}
    \caption{The `thinking time' of direct VLM control becomes a critical vulnerability in real-time games, highlighting the need for a paradigm shift on VLM use: from direct controller to execution module developer}
    \label{fig:intro1}
\end{figure}

Recent VLM-based approaches leverage VLMs as direct game controllers through a pause-and-plan paradigm \cite{tan2024cradle, chen2024can}: the agent periodically pauses gameplay, using VLM and vision tools (e.g., OCR, segmentation) together to analyze game screens, plan actions and then directly output key-mouse command to control game. However, this paradigm suffers from fundamental limitations: (1) it heavily depends on the game's support to pause at any moment, which disrupts the gameplay flow and limits its applicability to a narrow range of games that support such interruptions; (2) Requiring VLM reasoning for every action makes it unsuitable for tasks demanding high reactivity (e.g., FPS shooting); (3) VLM outputs simple key-mouse control commands without real-time interactive logic for game environments, making it hard solve tasks demanding dynamic adaptation (e.g., action game combat). These limitations reflect a \textbf{fundamental mismatch}: VLMs excel at time-consuming deliberate reasoning (scene understanding and planning) but struggle with rapid, continuous game interactions requiring millisecond-level responses (shown in Figure \ref{fig:intro1}). 

We observe that most human game actions rarely rely on deliberate reasoning, but rather flows from  quick-fire game sense - a set of trained reflexes and patterns developed through practice. This observation suggests a fundamental paradigm shift: Unlike using VLMs to directly control every game actions, we should elevate them to develop task-specific execution modules that can handle real-time interactions autonomously. These specialized modules, developed by VLM, solve specific tasks requiring rapid reactions or frequent environmental interactions. This paradigm shift bridges the VLM's reasoning with real-time gameplay demands, enabling more versatile game agents.

Based on this new paradigm, we present GameSense, a framework that empowers VLMs to develop and optimize task-specific execution modules, termed Game Sense Modules (GSMs). GameSense equips VLMs with essential tools, including vision tools and neural network training pipelines, to create GSMs tailored for diverse gameplay tasks. These modules can range from simple action-feedback loops (e.g., combat patterns based on HP bar monitoring) to complex, learned behaviors (e.g., boss fight strategies optimized through RL). These modules are seamlessly integrated into the gameplay loop: when the agent identifies a specific task, it activates the corresponding module and refines it based on execution feedback. By shifting VLMs' role from direct controller to the developer of GSMs, GameSense achieves efficient execution and promotes continuous improvement in gameplay performance.

Experiments demonstrate that GameSense is the first agent to achieve fluent gameplay in diverse game genres. In ACT/FPS games, our framework achieves the highest success rates in combat tasks, while achieving the highest exploration scores without gameplay pausing. In contrast, existing VLM-based methods either fail to complete such tasks or rely heavily on frequent gameplay pausing, disrupting the flow of real-time interactions. In the reflex-intensive game Flappy Bird where pausing is not supported, existing VLM-based methods fail at basic control, and GameSense develops precise control modules through iterative refinement. GameSense exhibits significantly improved real-time performance and adaptation capabilities, setting a new benchmark for game-playing agents.
%
%
%
The contributions of this paper are as follows:

\begin{itemize}
    \item We identify limitations of existing VLM-based game-playing approaches, particularly their inability to handle real-time, high-reactivity tasks.
    \item We propose a novel paradigm that uses VLMs to develop task-specific execution modules for autonomous real-time interactions.
    \item We introduce \textbf{GameSense}, a framework that enables VLMs to create and refine \textbf{Game Sense Modules (GSMs)}.
    \item Our experiments demonstrate that GameSense outperforms existing methods and is the first to master reflex-intensive games.
\end{itemize}

\section{Related Work}
\subsection{Environment for Video Gameplay and RL-based Agents}
Researchers have made significant strides in various video game environments, including classic games like Atari games\cite{bellemare2013arcade}, Minecraft\cite{fan2022minedojo,guss2019minerl}, StarCraft II\cite{ellis2023smacv2}. However, these environments rely heavily on open-source code or official APIs, requiring substantial human effort for implementation. This dependency restricts AI accessibility to general games. Recent RL-based approaches have attempted to overcome API dependencies by directly processing game visuals and simulating keyboard-mouse inputs, including DQN-play-sekiro\cite{myrepo}. However, these RL methods typically work for specific tasks and exhibit poor generalization, requiring retraining for new scenarios. 
The challenge of developing agents capable of generalizing across diverse gaming environments without API access remains largely unsolved. This limitation motivates our research toward a more adaptable solution using only visual inputs and key-mouse controls.

\subsection{LLM/VLM-Driven Gameplay Agent}
Current LLM/VLM-driven gameplay agents follow two main approaches. The first relies on game APIs for state observation and control, as seen in Minecraft\cite{wang2023voyager, liu2024rl} and Starcraft II agents\cite{ma2023large}. While effective, this API dependency limits their application to closed-source commercial games. The second approach uses VLMs to directly process screen information and generate keyboard-mouse controls, as demonstrated by Cradle\cite{tan2024cradle}. Though eliminating API requirements, this method's frame-by-frame analysis and decision-making process introduce significant latency. This makes such agents unsuitable for tasks requiring quick reactions or dynamic adaptation. While recent works like SIMA\cite{raad2024scaling} and VARP\cite{chen2024can} attempt to improve performance through behavior cloning, they require extensive human gameplay data for training. The challenge of achieving real-time and adaptive gameplay in VLM-driven agents remains unsolved, motivating our research toward a new paradigm.


\section{Method}
\begin{figure*}[t]
    \centering
    \includegraphics[width=\linewidth]{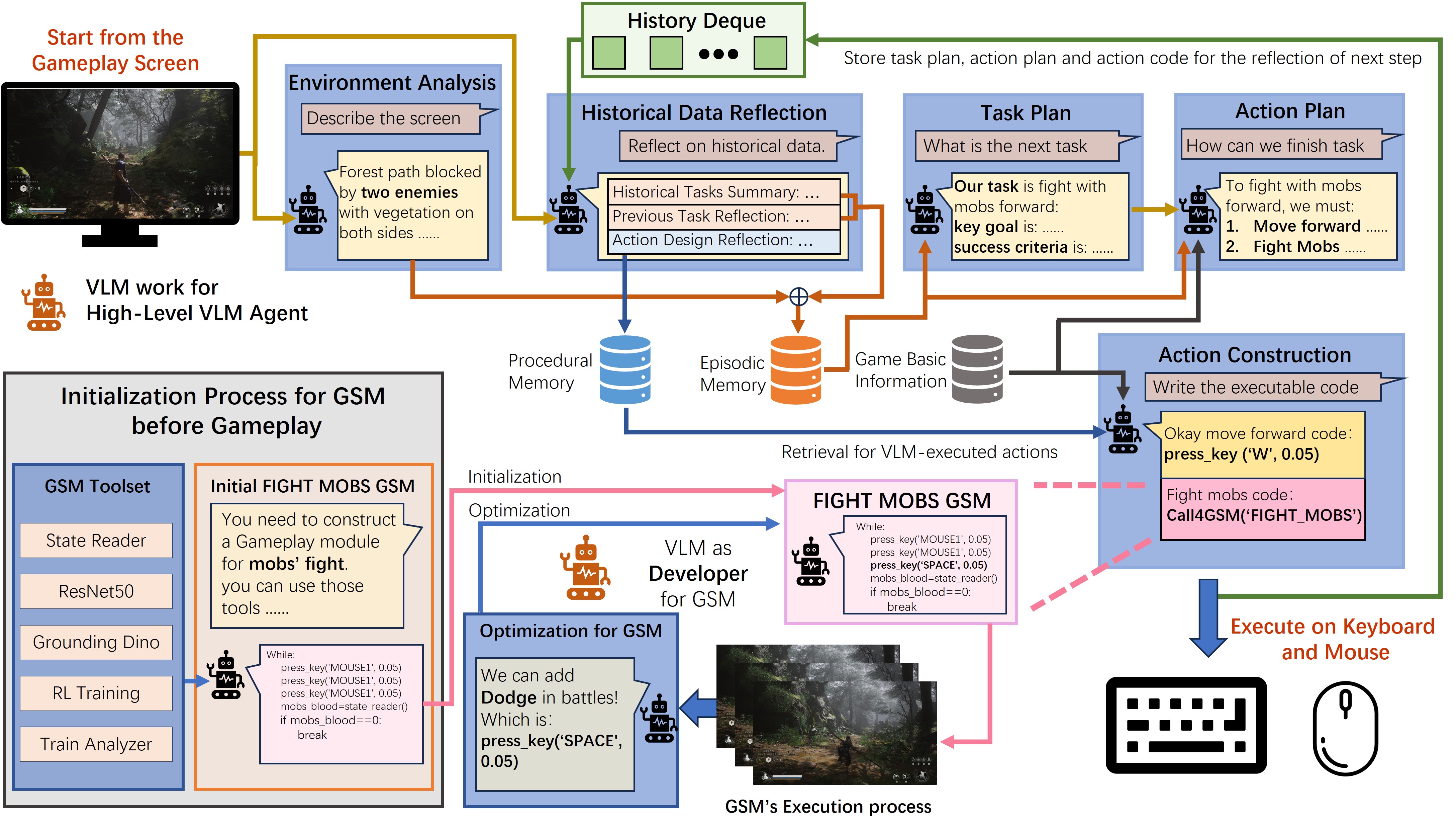}
    \caption{The overall architecture of GameSense. The main loop is governed by the VLM, which analyzes the game environment, reflects on history, plans tasks and actions, and constructs the code for each action (both VLM-executed and GSM actions). The VLM acts as a developer to refine GSMs through analysis GSM's execution process.}
    \label{fig:method1}
\end{figure*}

\subsection{Problem Formulation and Motivation}
This work aims to develop a real-time gameplay agent that operates \textbf{without} relying on game APIs or pausing the game for action reasoning. The agent solely depends on real-time game screens and outputs key-mouse control commands to interact with the game. This approach aims to create a truly \textbf{in-game} agent, mirroring how human players experience and interact with the game environment.

Existing gameplay agents rely on the "pause and plan for each action" paradigm, which exhibits limitations in fast-paced and dynamic game scenarios. In contrast, most human gameplay actions do not stem from deliberate reasoning over each move but from game sense—an intuitive ability to react swiftly based on experience.  Motivated by this observation, we propose an agent system capable of developing its form of "game sense," enabling more natural and efficient interaction in gameplay.

\subsection{Overview of GameSense}


GameSense introduces a \textbf{paradigm shift} by elevating the VLM from direct controller to developer of task-specific execution modules, termed Game Sense Modules (GSMs). The agent integrates a High-Level VLM Agent and GSMs: the High-Level VLM Agent is responsible for real-time game screen analysis, historical reflection, and task and action planning. The GSMs, independently developed by the VLM itself, handle tasks requiring rapid response (e.g., combat, shooting, rapid clicks). As shown in Figure~\ref{fig:method1}, the agent operates in a continuous loop: it analyzes real-time game screens, reflects on history, and plans tasks and actions. Depending on the action requirements, the agent either directly generates key-mouse control codes (\textbf{VLM-executed actions}) for straightforward actions or invokes GSMs (\textbf{GSM actions}) for high-speed processing. This process ensures efficient and natural interaction with the game, mirroring human-like gameplay.

\subsection{High-Level VLM Agent}
The High-Level VLM Agent serves as the brain of the system, responsible for understanding the game environment, reflecting on past experiences, and planning future tasks and actions (both VLM-executed and GSM actions). This module is structured into several core components:

\textbf{Game Environment Analysis:} This module leverages VLM's visual understanding capabilities to generate a textual description of the current game screens.  It identifies key elements such as the presence of enemies, bosses, interactable objects, potential threats, and the player character's status. This textual description is then used for historical reflection and task planning.
    
\textbf{Historical Data Reflection:} This module performs three parallel types of reflection to learn from the past: (1) Previous Task Reflection: evaluate the success of the previous task and suggesting optimizations; (2) Historical Task Summary: summarize the last 10 task executions to extract long-term patterns; and (3)  Action Design Reflection: assess \textbf{VLM-executed actions}’ effectiveness and generating refinements. This mechanism ensures continuous self-assessment and refinement.

\textbf{Memory:} This module serves as a structured repository for Historical Data Reflection and Game Environment Analysis, which consists of \textbf{episodic memory} and \textbf{procedural memory}. Episodic memory stores the Game Environment Analysis, Previous Task Reflection and Historical Task Summary, providing temporal context for the agent's understanding of game progression and task outcomes. This memory \textbf{directly} passed to the Task and Action Plan module, enabling the VLM to make context-aware decisions. Procedural memory, implemented as a RAG database, specializes in storing and retrieving action implementation experiences for \textbf{VLM-executed actions}. It stores action names, the corresponding action code, and the associated reflection results from Action Design Reflection. When planning a new VLM-executed action, the agent queries the procedural memory using the action name as the key, retrieving relevant historical data to guide action construction.

\textbf{Task Plan:} Based on the Episodic Memory, this module determines the next task the agent should undertake. It considers the overall current situation and past experiences to generate a high-level \textbf{task description}, including the key goal, success criteria, and locations (if needed).

\textbf{Action Plan:} Given the task description and the Episodic Memory, this module plans a sequence of action names required to complete the task. This planning is grounded in a predefined action mapping table that provides a comprehensive and conflict-free set of actions, including both \textbf{VLM-executed actions} (single key-mouse operation, e.g., "move forward": "use [key] to move") and \textbf{GSM actions} (calls to specialized GSMs, e.g., "Fight mobs": "invoke [Fight GSM] to fight mobs"). Each action in the table is accompanied by a clear textual description, enabling the VLM to leverage its language understanding capabilities to connect the task's semantic meaning with appropriate actions. For instance, when tasked with "engage the mobs ahead," the VLM references the mapping table to retrieve possible actions. By analyzing the action descriptions, the VLM constructs an ordered sequence of action names such as ["move forward" (VLM-executed), "Fight mobs" (GSM action)].

\textbf{Action Construction:} This module translates the planned action names into executable code, referencing the action mapping table and procedural memory. For VLM-executed actions, the VLM generates the key-mouse code (including both the specific key and its duration), leveraging the procedural memory for guidance. For GSM actions, this module simply outputs the code to call the appropriate GSM.

The High-Level VLM Agent operates in a closed-loop process. It begins by analyzing the game screen to understand the current state. Then, it reflects on past experiences through the three reflection mechanisms. Based on the current state and reflections, it plans the next task and the sequence of actions. Finally, it constructs the code for each action (both VLM-executed and GSM actions). This process is driven by the VLM's reasoning and code-generation capabilities,  with each cycle potentially contributing to improving future decision-making through memory and reflection. Further details are availabel in Appendix\ref{sec:A4h}.

\subsection{Game Sense Modules (GSMs)}

\subsubsection{Motivation and Design Philosophy}
Our goal is to achieve a “game sense” similar to that of human players—the ability to respond to gameplay dynamically, which is key to a successful real-time gaming experience. Specifically, we reposition VLM from a direct controller to a developer and optimizer, creating and continuously optimizing Game Sense Modules (GSMs). We require VLM design to follow a “from start to finish” design, which means each GSM is designed as a complete execution equipped with adaptive execution loops and termination criteria, rather than a mere sequence of actions. This design ensures both real-time performance and dynamic adaptability.

\subsubsection{GSM Types and Application Scope}
Our approach categorizes GSM into two types: (1) \textbf{RL-based GSM}, which is designed for scenarios requiring high dynamic adaptability where task patterns are difficult to model with fixed rules (e.g., boss fights and Flappy Bird control); (2) \textbf{Rule-based GSM} targets tasks with well-defined rules that demand rapid, efficient responses (e.g., mob fights and shooting in FPS games).

In each game, the tasks handled by GSMs are predefined during Agent initialization. In ACT games, GSMs handle mob fights and boss fights. In FPS games, GSMs manage shooting. In Flappy Bird, GSMs control the bird's flight. This design is based on the following reasons: (1) \textbf{Limited Game Sense Requirements:} For a specific type of game, a limited number of game sense modules are sufficient to support smooth gameplay (e.g., fight for ACT, shoot for FPS). (2) \textbf{Experimental Validation:} Experiments \ref{ab:fix} have shown that allowing the VLM to autonomously generate GSM modules is counterproductive. Excessive autonomy can lead to frequent and redundant GSM creation and low reusability of GSM, increasing computational overhead and management complexity. 

\subsubsection{GSM Toolset}
GSM relies on the following general-purpose tools for task execution. We argue that the use of such tools is well-justified: (1) it mimics humans' direct understanding of game visuals; (2) existing methods \cite{tan2024cradle, liu2024rl} commonly depend on general-purpose visual tools. 

The key tools include: (1) \textbf{State Reader:} An OpenCV-based game frame analyzer for extracting game states (e.g., HP bars, death status).(2) \textbf{Vision Processors:} Including ResNet50 \cite{7780459} or CNN for feature extraction and Grounding Dino \cite{liu2024grounding} for object detection. These are standard computer vision models. (3) \textbf{RL Training Parent Class:} A standard RL training parent class implementation for building RL-based GSMs, which requires VLM to instantiate it. (4) \textbf{Training Analyzer:} For analyzing training process data, including reward curves and behavior statistics, providing optimization insights for VLM. Further details of toolset and case presentation are available in Appendix \ref{ap:tool}.

These tools are standard components in computer vision and RL.  The key innovation of GSM innovation lies in how VLM develops GSMs rather than the tools themselves.

\subsubsection{RL-based and Rule-based GSMs}
\textbf{RL-based GSM} designed for tasks requiring dynamic adaptation (e.g., boss fights, Flappy Bird). VLM firstly designs the state space (by selecting relevant  states from the output of \textbf{State Reader}, like HP state of character/boss), action space (by selecting task-relevant controls from key-mouse mappings) and constructs initial reward functions based on task objectives. Based on the above, \textbf{RL Training Parent Class} is instantiated, and then RL training is initiated. As training begins, VLM optimizes reward function through \textbf{Training Analyzer}. This process establishes a "train-analyze-optimize" loop,  enabling GSM to progressively master complex task execution strategies.

\textbf{Rule-based GSM} focuses on tasks with clear logic but demanding quick reactions (e.g., FPS shooting, mob fights). During creation, VLM first analyzes task objectives and selects necessary visual processing tools (e.g., Grounding Dino for shooting), then designs a complete control loop with execution logic and end conditions. During execution, VLM optimizes the execution logic through screen analysis, such as adjusting the Grounding Dino label list for more precise shooting target detection. This "execute-analyze-optimize" loop ensures GSM maintains continuously improved execution precision. 

Both GSM approaches have a “from start to finish” design. And we suggest setting the max optimization iterations of GSMs to 3 (Show in \ref{opt_num}). Further details are available in Appendix \ref{ap:rl}.

\subsection{System Integration}
Before the agent begins gameplay, the system is initialized with the following components: (1) Game Mechanics and Objectives: A detailed description of the game mechanics, including rules, objectives, and success criteria; (2) Predefined action mapping table: serves as the foundation for agent-game interaction, containing both basic key-mouse control mappings and predefined GSM action, each with detailed functional descriptions; (3) GSM Module Initialization: initialization based on the predefined GSM actions' description and tool instructions. RL-based GSM initializes action space, state space, and reward function. Rule-based GSM initializes execution logic and end conditions. Then the agent operates in a continuous loop (High-Level VLM Agent) and the GSM module continuously optimizes its performance in a parallel process.

\section{Experiment}

\subsection{Implementation Details}
To ensure reproducibility, we adopt an open-source VLM with Qwen 2.5 VL as the backbone. All games are run on a single Windows machine equipped with an NVIDIA 4060 GPU. This setup guarantees that the experimental results can be reliably reproduced and provides a clear reference for the hardware environment used in our evaluations.

\subsection{Evaluation Methods}
Our evaluation focuses on two aspects: (1)\textbf{Single-Task Performance:} We select important tasks within each game to assess the agent's task completion rate. For instance, in the ACT game (e.g., combat with minor monsters and boss battles), in the FPS game (e.g., shooting and movement), we evaluate how effectively the agent handles these critical tasks that demand high real-time responsiveness. (2) \textbf{Complete Game Flow Evaluation:} We let all agents independently engage with and adapt to the game using a fixed initial scenario. The evaluation metrics include max exploration scores (how comprehensively the agent navigates the environment) and the average exploration scores, which validate the agent's overall gameplay capabilities. 

\subsection{Baselines}
We compare our approach, GameSense, with Cradle—the only general game agent specifically designed for video games \cite{tan2024cradle}. For a comprehensive comparison, we evaluate both the standard Cradle and its variant without the stop mechanism (Cradle without stop). It is important to note that GameSense \textbf{does not require any pausing}, thereby offering significant advantages in real-time performance and seamless gameplay.

\subsection{Result of Single-Task}

\begin{table*}[h]
\centering
\resizebox{\textwidth}{!}{%
\begin{tabular}{|l|c|c|c|c|c|c|}
\hline
\multicolumn{7}{|c|}{\textbf{Black Myth: Wukong (not support an immediate pause during combat or under attack)}} \\
\hline
& UI Operation & Map Escape & Item Interaction & Normal Mob Battle& Harder Mob Battle & Boss Battle \\
\hline
Cradle & 95\% & 55\% & \textbf{75\%} & 25\% & 10\% & 0\% \\
\hline
Cradle w/o stop& 95\% & 50\% & 70\% & 0\% & 0\% & 0\% \\
\hline
GameSense & \textbf{100\%} & \textbf{60\% }& 70\% & \textbf{95\%} & \textbf{70\%} &\textbf{ 60\%} \\
\hline
\end{tabular}
}

\vspace{1em}

\resizebox{\textwidth}{!}{%
\begin{tabular}{|l|c|c|c|c|c|}
\hline
\multicolumn{6}{|c|}{\textbf{DOOM (supports pausing at any moment)}} \\
\hline
& UI Operation & Map Escape & Interact with Door & Normal Mob Shot & Harder Mob Shot \\
\hline
Cradle & 95\% & 45\% & 35\% & 10\% & 5\% \\
\hline
Cradle w/o stop & \textbf{100\%} & 30\% & 35\% & 0\% & 0\% \\
\hline
GameSense & 95\% & \textbf{50\%} & \textbf{40\%} & \textbf{85\% }& \textbf{65\%} \\
\hline
\end{tabular}
}
\caption{Single-task experiment on Black Myth: Wukong and DOOM. Before testing, we let Cradle run 10 steps in specific scenarios to adapt to the situation. For GameSense, we run 10 steps in specific scenarios and optimize the GSM through three iterations. For GSMs, Mob battle/shot corresponds to rule-based GSMs and Boss battle corresponds to RL-based GSMs.} 
\label{tab:single_task}
\end{table*}

In our experiments on the ACT game ``Black Myth: Wukong'', the following tasks were defined: (1) \textbf{UI Operation:} Using the in-game UI to restore blood volume. (2) \textbf{Map Escape:} Resolving issues where the character gets stuck at the map boundary, by adjusting the camera view. (3) \textbf{Approach to Item Interaction:} Moving close to the shrine for interaction. (4) \textbf{Normal Mob Battle:} A combat task where a monster can be defeated with three hits. (5) \textbf{Harder Mob Battle:} A more challenging combat task requiring six or seven hits. (6) \textbf{Boss Battle:} A high-difficulty combat task. In our experiments on the FPS game ``DOOM'', the following tasks were defined: (1) \textbf{UI Operation:} Using the UI to enter the game. (2) \textbf{Map Escape:} Make the character turn correctly at the right angle of the road, by adjusting the camera view. (3) \textbf{Interact with Door:} Moving close to the interactive door and open it. (4) \textbf{Normal Mob Battle:} A shot task where the monster has slow movement speed. (5) \textbf{Harder Mob Battle:} A more challenging shot task where the monster has fast movement speed. The experiment for each task was repeated 20 times.


\textbf{Note on Pause Mechanism:} Black Myth: Wukong does not support an immediate pause during combat or under attack. To run Cradle, we had to implement a mechanism where a pause is attempted up to 5 times; if pausing still fails, the system abandons the pause. This increases the risk of the character being attacked during the VLM’s reasoning, highlighting a significant compatibility issue with Cradle. DOOM supports pausing at any moment, which enables Cradle to run normally.

Table~\ref{tab:single_task} summarizes the success rates for each task. In non-real-time tasks, all three methods demonstrated similar performance (typically ranging from 50\% to 95\%). However, Cradle (without stop) showed a significant decrease to 30\% in DOOM's map escape task due to potential unexpected monster encounters, where its inferior reaction capability renders it completely ineffective. In combat scenarios, GameSense demonstrated overwhelming superiority, achieving success rates of 60\%-95\% in Black Myth: Wukong and 65\%-85\% in DOOM, while other methods were practically unusable in combat situations (with success rates of only 0-25\%). These results convincingly demonstrate the exceptional capabilities of the GameSense framework in handling complex real-time interaction scenarios.

\subsubsection{Result of Complete Game Flow}
\begin{figure*}[t]
    \centering
    \includegraphics[width=1\linewidth]{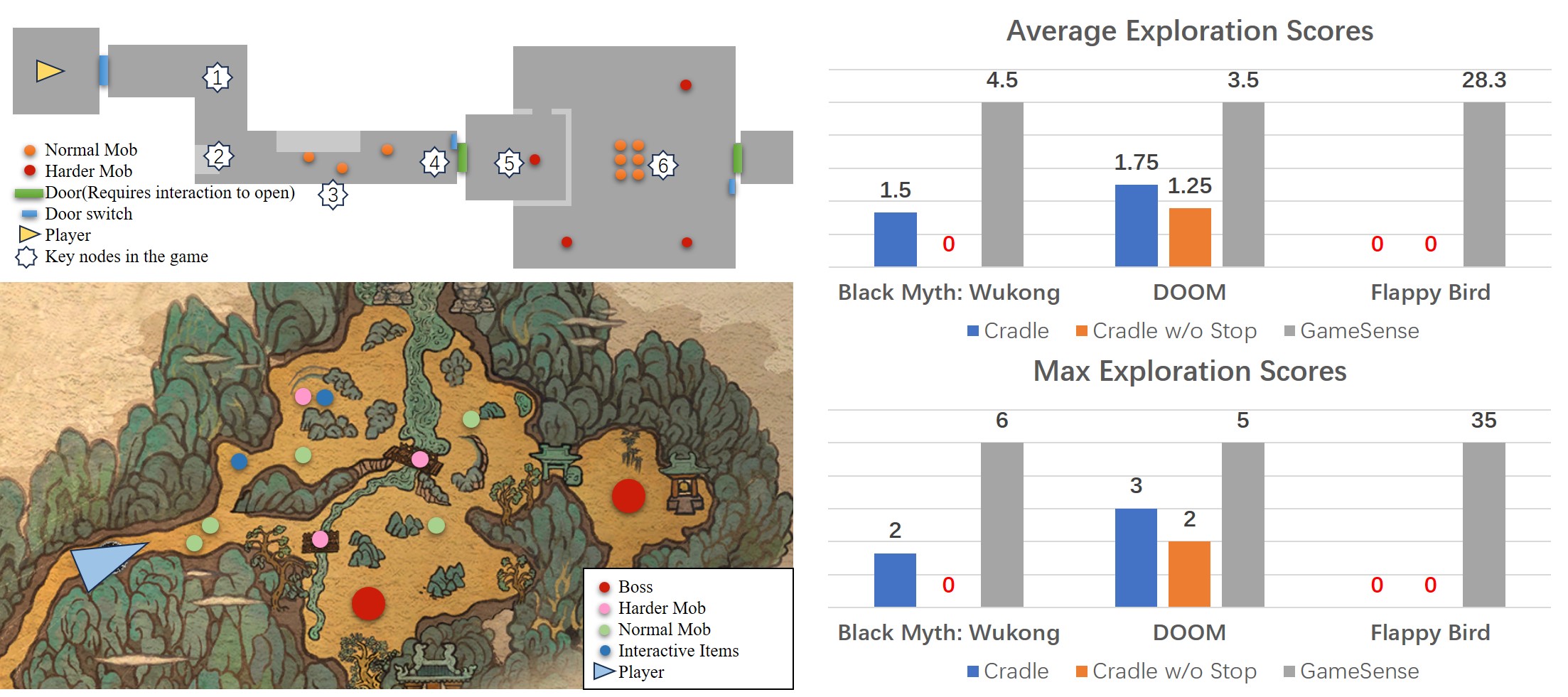}
    \caption{Complete game flow performance on Black Myth: Wukong, DOOM, and Flappy Bird. }
    \label{fig:exp1}
\end{figure*}
Map of "Black Myth: Wukong" and "DOOM", as shown in Figure \ref{fig:exp1}. To evaluate the complete game flow, we use the exploration progress in games as a performance metric, with different criteria defined for each game. For "Black Myth: Wukong," considering its open-world map, we score based on the consecutive tasks completed by the Agent: defeating a normal mob scores 1 point, successfully navigating a junction scores 1 point, defeating a harder mob scores 2 points, and successful interaction with items (such as collecting herbs or treasures) scores 1 point. For "DOOM," given its linear map, we have marked key points on the map, including turning, shooting enemies, and interacting with doors, with each key point passed scoring 1 point. For "Flappy Bird," we measure how many pipes the bird passes, with each pipe scoring 1 point. For all games, we calculate the total score from the starting point to the character's death. In our experimental setup, each game was run 20 times from a fixed initial position, and two primary metrics were recorded: the average number of explored scores and the maximum score achieved by the agent.



As shown in figure \ref{fig:exp1}, the experimental results clearly demonstrate the superior performance of GameSense in-game exploration tasks: in the open-world ACT game "Black Myth: Wukong," it achieved an average exploration score of 4.5 and a maximum score of 6.0; in the linear level game "DOOM," it reached an average score of 3.5 and a maximum score of 5.0; and in the continuous reaction game "Flappy Bird," it impressively scored an average of 28.3 and a maximum of 35. In contrast, Cradle performed poorly or failed to effectively play the games at all, strongly validating GameSense's significant advantages in achieving authentic gameplay experiences and its versatility across different game genres.
\subsection{Ablation Study}
\subsubsection{RL-based GSM}

Although our RL-based GSM utilizes a general-purpose RL Training Parent Class rather than one specifically tailored for individual game scenarios, the stringent requirements for training RL models still make it challenging to establish complete training protocols across all gaming environments. This raised concerns about whether Rule-based GSM alone could enhance agent capabilities, when RL training is prohibited. Therefore, we conducted experiments in boss battle scenarios, where VLM solely develop Rule-based GSM. As shown in Table\ref{tab:RL GSM}, Rule-based GSM still managed to reduce the boss's health to 34.6\% and achieve a success rate of 10\%. These results indicate that rule-based GSM also significantly enhance the Agent's combat capabilities. Furthermore, this indicates that our \textbf{paradigm shift}, which transforms VLM from a direct controller to a GSM observer, is the \textbf{key} to enhancing agent capabilities. Detailed analysis and experiment setting can be seen in appendix \ref{ap:rlgsm}.
\begin{table}[h]
  \centering
  \begin{tabular}{|l|c|c|}
    \hline
    GSM Type & Success Rate & Avg Blood \\
    \hline
    RL-based   & 60\% & 12.3\% \\
    Rule-based      & 15\% & 34.6\% \\
    Cradle     & 0\% & 90.2\% \\
    Cradle w/o Stop & 0\%   & 95.8\% \\

    \hline
  \end{tabular}
  \caption{Avg Blood means the average remaining health of the boss, which also represents the combat ability of different agents.}
    \label{tab:RL GSM}
\end{table}

\begin{table}[h]
  \centering
  \resizebox{0.49\textwidth}{!}{
  \begin{tabular}{|l|c|c|c|}
    \hline
        \multirow{2}{*}{Opt Num} & \multicolumn{2}{c|}{Battle Tasks in ACT Game} & Flappy Bird \\
    \cline{2-4}
    & Success Rate (Mobs) & Success Rate (Boss) & Pipes Passed \\
    \hline
    0 & 70\% & 10\% & 18.3 \\
    1 & 90\% & 40\% & 28.1 \\
    2 & 100\% & 50\% & 27.3 \\
    3 & 90\% & 60\% & 28.2 \\
    \hline
  \end{tabular}
  }
  \caption{Impact of GSM optimization iterations on gaming tasks. The first two columns show task success rates as percentages, while the last column indicates the average number of pipes passed in Flappy Bird.}
  \label{tab:opt}
\end{table}

\subsubsection{Optimization Iterations of GSM}
\label{opt_num}
We investigated the impact of GSM optimization iterations on its performance by extracting multiple iterative versions of GSM and testing their performance. As shown in table\ref{tab:opt}, while one to two optimization iterations are sufficient for simpler tasks, more complex challenges like boss battles benefit from additional optimization cycles, highlighting the importance of iterative refinement in GSM's performance. Detailed analysis and experiment setting can be seen in appendix \ref{ap:optGSM}.

Additionally, we found that there is a certain probability of degradation occurring when the number of GSM optimizations is too high. This is due to the accumulation during the optimization process, with more bad cases and optimization case-by-case analysis as shown in the appendix\ref{ap:case}. So we suggest setting the maximum number of iterations for optimization to 3.


\subsubsection{Unfixed GSM}
\label{ab:fix}
Although we have emphasized that the fixed GSMs are sufficient for specific gaming scenarios, we remain concerned about whether allowing the VLM to autonomously develop GSMs could broaden their applicability. Therefore, we integrated an additional step in the high-level VLM agent, permitting the VLM to independently reason about and design GSMs. Unfortunately, we observed that the GSMs autonomously generated by the VLM were often repetitive, with the VLM designing duplicate GSMs for each encountered mob. This frequent construction of GSMs not only places extra operational demands on the Agent but also necessitates prolonged decision-making times, compelling us to pause the game frequently, contrary to our initial objectives. Appendix \ref{sec:A4c} shows more details.

\section{Conclusion}
In this paper, we first identify a common issue with existing VLM-based gameplay agents: the VLM infers each action individually, resulting in significant "thinking delays", which limits their capability to handle real-time and dynamically adaptive tasks. To address this issue, we propose a paradigm shift, transforming the VLM's role from a direct controller to a developer of game action execution modules. Furthermore, we developed the GameSense, which is the first agent capable of performing tasks such as shooting in FPS games and boss fights in ACT games without game's pause function. This provides a new paradigm for construct VLM-based gameplay agents.

\section{Limitation}
This paper introduces a paradigm shift in the design of VLM gameplay agents: transforming VLMs from direct action controllers to developers of Game Sense Modules (GSMs). Although our experiments have proven the effectiveness of this approach, there remains an issue. For each game, the types and functions of GSMs are fixed. While we have discussed that this fixed nature is sufficient for gameplay and that complete autonomy in design by the VLM would introduce catastrophic delays, exploring how to enable VLMs to autonomously recognize and reuse GSMs is still worthwhile, as it could broaden the applicability of Gameplay Agents.



\bibliography{custom}
\clearpage
\appendix
\section{Detail for High-Level VLM Agent}
\label{sec:A4h}
\subsection{Detailed Input/Output for Each Module}

1. \textbf{Game Environment Analysis}
\begin{itemize}
    \item \textbf{Input:} Real-time game screen images captured directly from the game.
    \item \textbf{Output:} A detailed textual description that identifies key elements in the scene—such as enemies, bosses, interactable objects, potential threats, and the current state of the player.
\end{itemize}

2.1 \textbf{Previous Task Reflection} (Historical Data Reflection)
\begin{itemize}
    \item \textbf{Input:} Data from the most recent task execution (screenshots), description of the previous task, and action design. description.
    \item \textbf{Output:}  A detailed evaluation of the latest task's performance that highlights immediate strengths, weaknesses, and suggestions for the next task design.
\end{itemize}

2.2 \textbf{Historical Task Summary} (Historical Data Reflection)
\begin{itemize}
    \item \textbf{Input:} Aggregated data from a sliding window of recent tasks (e.g., the last 10 tasks), including task description, logs, and task reflection.
    \item \textbf{Output:} A synthesized summary that identifies long-term trends, and recurring patterns, providing broader context for decision-making.
\end{itemize}

2.3 \textbf{Action Design Reflection} (Historical Data Reflection)
\begin{itemize}
    \item \textbf{Input:} Data related to VLM-generated action executions including screenshots, design of task and action.
    \item \textbf{Output:}  A detailed evaluation of the action design of the latest task that highlights immediate strengths, weaknesses, and suggestions for optimization.
\end{itemize}

3.\textbf{ Memory:}
\begin{itemize}
    \item Input: Reflection outputs from the Historical Data Reflection module.
    \item Output: Two types of stored memory:
    \begin{itemize}
    \item Episodic Memory: Time-indexed records of past task outcomes (both Previous Task Reflection and Historical Task Summary).
    \item Procedural Memory: A RAG database mapping action names to their corresponding key-mouse control codes and associated reflection data. 
\end{itemize}
\end{itemize}

4. \textbf{Task Planning:}
\begin{itemize}
    \item \textbf{Input:} The textual description from Game Environment Analysis along with contextual insights from Episodic Memory.
    \item \textbf{Output:} A high-level task description that specifies the core objective, success criteria, and any relevant spatial or situational details for the current game scenario.
\end{itemize}

5. \textbf{Action Planning:}
\begin{itemize}
    \item \textbf{Input:} The high-level task description generated by Task Planning.
    \item \textbf{Output:} An ordered list of action names derived from a predefined action mapping table.
\end{itemize}

6. \textbf{Action Construction:}
\begin{itemize}
    \item \textbf{Input:} The ordered list of action names from Action Planning, along with reference data from Procedural Memory and the action mapping table.
    \item \textbf{Output:} Executable control codes that translate into either detailed key-mouse commands (for VLM-executed actions) or invocation instructions that trigger the corresponding Game Sense Modules (for GSM actions), enabling real-time game control.
\end{itemize}

\subsection{Implementation Details of FPS Game}
FPS games have a unique mechanism where attacks are primarily executed through shooting. This means that players can open fire as soon as they spot an enemy, and similarly, enemies will shoot upon detecting the player. To cater to the game's demand for shooting at any moment, we have automated the invocation of the Shooting GSM after each module in the high-level VLM agent for FPS games, significantly reducing the risk of the agent being attacked by enemies. During the design process of the GSMs by the VLM, termination and exit mechanisms were also considered. For instance, if the Grounding Dino fails to detect enemies multiple times, it will exit the Shooting GSM, ensuring that this mechanism does not interfere with other processes of the high-level VLM agent. Additionally, the Action Planning module is still allowed to invoke the Shooting GSM to handle a variety of game scenarios.

\section{Detail for Game Sense Modules}
\label{sec:A4S}
\subsection{Detailed Introduce for Part of Tool Set}
\label{ap:tool}
\textbf{The RL Training Parent Class} is a universal RL training class that defines a complete RL training workflow skeleton. At its core is the QNetwork neural architecture, which employs a triple-branch parallel processing design: a vision model (normal CNN for Flappy Bird, Resnet50 for ACT Boss Battle) branch for visual feature processing, a state branch for state information processing, and an action history branch using LSTM for processing historical action sequences. These three branches ultimately merge their features for decision-making, making it particularly suitable for handling complex state spaces and action sequences in video games.

The parent class includes the DoubleDQN Training Module, which implements core DoubleDQN algorithm functionalities, featuring experience replay memory, exploration strategy, and soft target network updates. The parent class also provides interfaces for model saving and loading, supporting training interruption and resumption. The training process is uniformly managed by the \textbf{train() } method, supporting multiple training episodes, with each episode executing standard operations such as environment interaction, experience collection, parameter updates, and training log recording.

To utilize this training parent class, specific scene subclasses need to be instantiated through VLM, primarily customizing \textbf{state space, action space, and reward functions}. Once the subclass is instantiated, training can be initiated directly using the parent class's \textbf{train()} method. During training, the framework automatically manages model checkpoint saving and training log recording. Through VLMs overriding of the reward function method, reward strategies can be flexibly adjusted. This design pattern allows VLM to focus on strategy optimization for specific games while reusing standard training workflows, making it applicable to various video games requiring visual input and continuous action decision-making.


\textbf{Training Analyzer} analyzes the training record data generated during the RL training process. Its purpose is to analyze and compile training statistics, which are then submitted to the VLM to assess whether the RL training meets expectations and optimize the reward accordingly. The module analyzes character state data (including health, mana, stamina, etc.) and action data, calculates key metrics such as total training steps, average rewards, and action usage frequency, and generates visualization charts including cumulative reward curves and state variable trends. These comprehensive statistical results enable the VLM to evaluate the model's training effectiveness and optimize the reward design accordingly. Based on these comprehensive statistical results, VLM can evaluate whether the RL model has learned to use various actions reasonably, whether the training process is stable, whether it has achieved the expected game goals (such as reducing Boss health), and whether the reward design is reasonable.

\subsection{Detailed Pipeline for RL-base GSM}
\label{ap:rl}
\subsubsection{Overview}
The RL Training Parent Class can be instantiated by VLM through a systematic process tailored to different game environments. The implementation consists of several key components and processes.

The RL Training Parent Class can be instantiated by VLM through a systematic process tailored to different game environments. First, we provide an RL training environment restart functionality to VLM. For ACT games, we leverage in-game teleportation cheats to enable precise character repositioning after respawn. For Flappy Bird, where revival requires a simple click, we implement a game-over detection module.

In instantiating the RL Training Parent Class, VLM employs the state reader to design the state space (e.g., character/boss status) and action space. Based on task objectives, VLM constructs an initial reward function. For example, in ACT games, the state space might include character health, boss health percentage, and relative positions, while in Flappy Bird, it might track bird height and scores achieved.

As training commences, VLM utilizes its Training Analyzer to optimize the reward function. This creates a "train-analyze-optimize" loop where VLM: (1) Monitors agent performance through training logs; (2) Adjusts reward signals to encourage desired behaviors; 
(3) Updates the reward function implementation. 

This iterative process enables GSM to progressively master complex task execution strategies, adapting to different game scenarios while maintaining the fundamental training structure defined in the parent class. The flexibility of this approach allows for continuous refinement of the training process while ensuring consistency in the underlying RL framework.

\subsubsection{Details of Initialization}
\textbf{For the state space, }we provide the VLM game task description (e.g. your task is to defeat the boss in the scene) and the \textbf{State Reader}. VLM selects task-related states to form a state space. This state space would be used to design the reward function.
\begin{tcolorbox}[breakable,colback=blue!5!white, colframe=blue!75!black, title=Example of State Space Design]
 Boss Blood (idx: 0)\\
Player Blood (idx: 1)\\
Potion Percentage (idx: 2)
\end{tcolorbox}

\textbf{For the action space,} we provide the VLM game task description and the game's action and key mode mapping table. VLM selects task-related actions from the mapping table to form an action space.
\begin{tcolorbox}[breakable,colback=blue!5!white, colframe=blue!75!black, title=Example of Action Space Design]
\begin{itemize}
    \item Move Forward (idx: 0)\\
    Basic movement action, no resource consumption or attack behavior involved.
    \item Move Backward (idx: 1)\\
    Basic movement action, no resource consumption or attack behavior involved.
    \item Move Left (idx: 2)\\
    Basic movement action, no resource consumption or attack behavior involved.
    \item Move Right (idx: 3)\\
    Basic movement action, no resource consumption or attack behavior involved
    \item Light Attack (idx: 4)\\
    Light attack deals damage to the Boss but consumes some stamina.
    \item Heavy Attack (idx: 5)\\
    Heavy attack requires charging time and can be interrupted, but deals higher damage. Best used when opportunity arises.
    \item Dodge (idx: 6)\\
    Dodge is used to avoid attacks, preventing HP loss when successful, but consumes stamina.
    \item Drink Health Potion (idx: 7)\\
    Drinking potion recovers HP but consumes potion stock. Suitable to use when HP is low.
    \item Cast Body Fixing (idx: 8)\\
    Casting immobilization spell requires mana, can control the Boss for a period of time, creating opportunity for damage output.
\end{itemize}
\end{tcolorbox}


For the initial reward function, we provide the VLM game task description, the state space, the action space, and Reward Function Template (standardizes input and output to ensure correct invocation by RL training classes, providing basic design ideas). Then, VLM independently designed reward function.

\textbf{Reward Function Template}:
\begin{lstlisting}[
    language=Python,
    breaklines=true,
    basicstyle=\ttfamily\small,
    keywordstyle=\color{blue},
    commentstyle=\color{green!60!black},
    stringstyle=\color{purple},
    showstringspaces=false,
    numbers=left,
    numberstyle=\tiny\color{gray},
    frame=none
]
def reward_function(prev_state, next_state, action_idx, done, action_history, action_state_changes, episode_start_time, step_time, step):
    # Initialize reward
    reward = 0.0

    # Game over logic
    if done:
        # Reward based on boss health reduction
        boss_health_reduction = 1-prev_state["boss_percentage"]
        # Design your reward logic
        ......
        return reward

    # Boss health change reward; Suggest giving linear rewards
    boss_health_change = prev_state["boss_percentage"] - next_state["boss_percentage"]
    # Design your reward logic
    ......
    reward+ = ......

    # Player health change reward; Suggest giving linear rewards
    player_health_change = next_state["blood_percentage"] - prev_state["blood_percentage"] 
    # Design your reward logic
    reward -= ......


    # Dodge-specific reward
    action = action_state_changes[action_idx]
    if action["action_name"] == "Dodge":
        # Design your reward logic
        ......
        reward+ = ......

    # Combo rewards/penalties
    def calculate_combo_reward(action_history):
        combo_reward = 0
        # Reward for consecutive light attacks

        if action_history[-4:] == ......:
        # Design your reward logic
            combo_reward += ......  
        # Penalty for excessive dodging
        if action_history.count(6) ......:
            # Design your reward logic
            combo_reward += ...... 
        # Penalty for excessive potion use
        if action_history.count(7) .....:
            # Design your reward logic
            combo_reward += ...... 
        return combo_reward

    reward += calculate_combo_reward(action_history)

    return reward
\end{lstlisting}

\textbf{Example of reward function} designed by VLM
\begin{lstlisting}[
    language=Python,
    breaklines=true,
    basicstyle=\ttfamily\small,
    keywordstyle=\color{blue},
    commentstyle=\color{green!60!black},
    stringstyle=\color{purple},
    showstringspaces=false,
    numbers=left,
    numberstyle=\tiny\color{gray},
    frame=none
]
def reward_function(prev_state, next_state, action_idx, done, action_history, action_state_changes, episode_start_time, step_time, step):
    # Initialize reward
    reward = 0.0

    # Game over logic
    if done:
        # Reward based on boss health reduction
        boss_health_reduction = 1-prev_state["boss_percentage"] 
        if boss_health_reduction >= 0.5:
            reward += 150  # Major reduction bonus
        elif boss_health_reduction >= 0.2:
            reward += 75   # Medium reduction bonus
        elif boss_health_reduction >= 0.1:
            reward += 30   # Minor reduction bonus
        else:
            reward -= 5   # Penalty for insignificant reduction
        return reward

    # Boss health change reward
    boss_health_change = prev_state["boss_percentage"] - next_state["boss_percentage"]
    if boss_health_change > 0.02:
        reward += 100 * boss_health_change 
    else:
        reward -= 2  

    # Player health change reward
    player_health_change = next_state["blood_percentage"] - prev_state["blood_percentage"] 
        reward += 10 * player_health_change


    # Dodge-specific reward
    action = action_state_changes[action_idx]
    if action["action_name"] == "Dodge":
        if player_health_change == 0:
            reward += 2  
        else:
            reward -= 0.5

    # Combo rewards/penalties
    def calculate_combo_reward(action_history):
        combo_reward = 0
        # Reward for 4 consecutive light attacks
        if action_history[-4:] == [4, 4, 4, 4]:
            combo_reward += 5  
        # Penalty for excessive dodging
        if action_history.count(6) > 15:
            combo_reward -= 5 
        # Penalty for excessive potion use
        if action_history.count(7) > 3:
            combo_reward -= 5 
        return combo_reward

    reward += calculate_combo_reward(action_history)

    return reward
\end{lstlisting}


\begin{figure*}[t]
    \centering
    \includegraphics[width=1\linewidth]{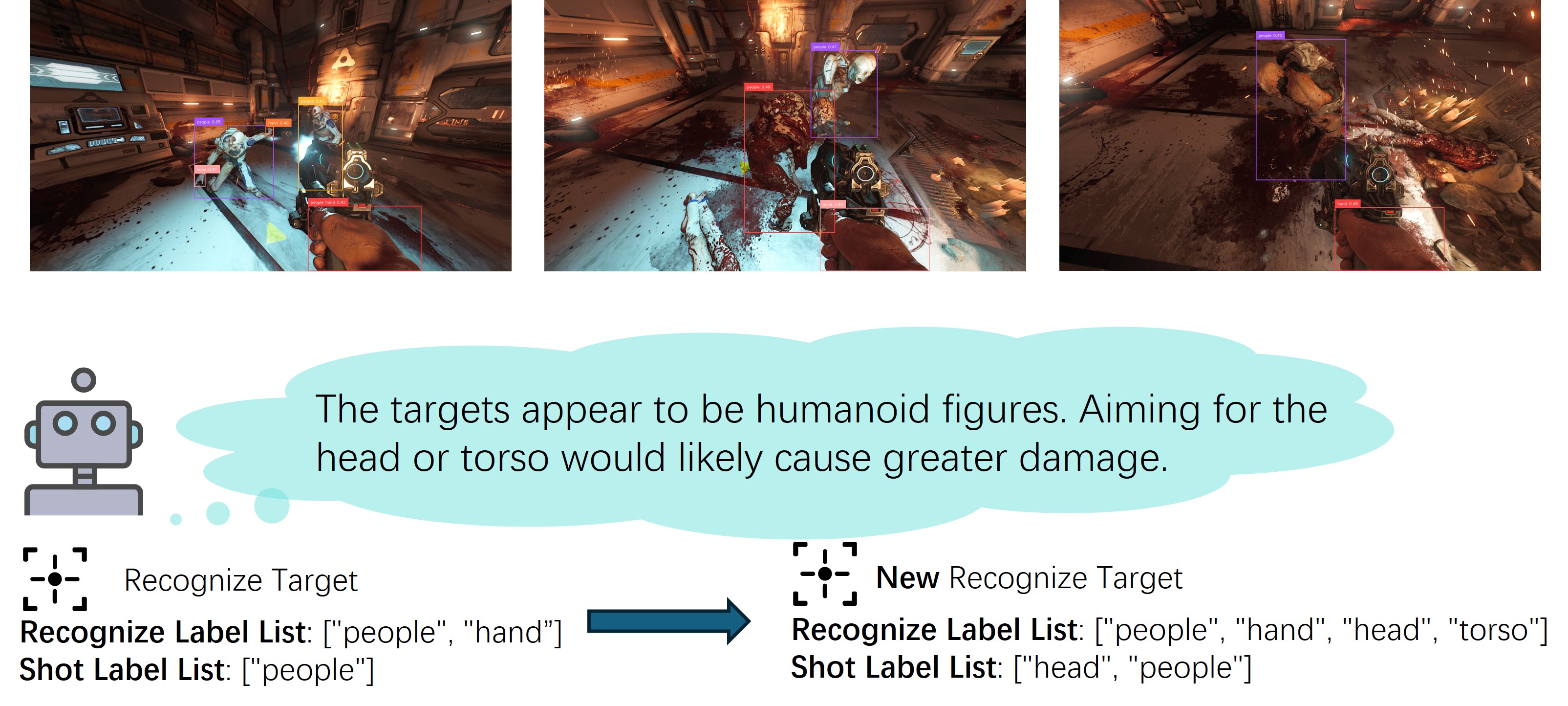}
    \caption{Case analysis for GSM's Optimization. }
    \label{fig:ap1}
\end{figure*}

\section{Detial of Ablation Study}
\label{sec:A4c}

\subsection{RL-based GSM}
\label{ap:rlgsm}
Both Rule-based and RL-based GSM underwent three iterations of optimization and the experiment for each GSM was \textbf{repeated 20 times}. Rule-based GSM still managed to reduce the boss's health to 34.6\% and achieve a success rate of 10\%. In contrast, Cradle completely failed to achieve any victories (zero success rate) and could barely inflict meaningful damage to the boss (remaining health at 90.2\% and 95.8\% respectively). These results indicate that rule-based GSM also significantly enhance the Agent's combat capabilities.

\subsection{Optimization Iterations of GSM}
\label{ap:optGSM}
Each GSM version was tested 10 times. The experimental results are shown in table\ref{tab:opt}. Starting from the unoptimized version (0 iterations), each optimization step generally improved performance until reaching optimal levels. These results indicate that while one to two optimization iterations are sufficient for simpler tasks like normal mob battles, more complex challenges like boss battles benefit from additional optimization cycles, highlighting the importance of iterative refinement in GSM's performance. 

\subsection{Unfixed GSM}
\label{ap:fix}

We incorporated an additional step in the high-level VLM agent, enabling it to independently conceptualize and develop GSMs. However, we observed that the GSMs spontaneously created by the VLM exhibited significant repetition, often designing duplicate GSMs for each encountered mob. This redundancy severely undermines the reusability of the GSMs, leading to the production of numerous low-quality, unoptimized GSMs. Table \ref{tab:apx} has shown this phenomenon.

\begin{table}[h]
  \centering
  \begin{tabular}{|l|c|c|}
    \hline
     & Num of GSM & Avg Opt \\
    \hline
    Unfixed GSM  & 12 & 0.17 \\
    fixed GSM & 2   & 3(Max) \\
    \hline
  \end{tabular}
  \caption{Avg Opt means average optimization iterations of GSM. We set the max optimization iterations to 3.}
    \label{tab:apx}
\end{table}

\subsection{Case-by-case Analysis for GSM's Optimization}
\label{ap:case}

The figure \ref{fig:ap1} demonstrates how the VLM optimizes the shooting GSM. The shooting GSM is designed based on the target detection capabilities of Grounding Dino, and thus the labels input by Grounding Dino directly impact performance. Initially, the VLM could only generate broad labels such as "people" and "hand." However, after observing the images detected during the execution process, the VLM enriched the list of labels, leading to performance optimization.


The following code example shows a reward optimization case. VLM found through analysis of training data that the proportion of dodge usage is too high, which is due to the excessive reward value for dodge behavior. This will cause the player to frequently dodge without attacking, so VLM has lowered the reward for dodging behavior and lowered the threshold for frequent dodging punishment. 

\textbf{Code before optimization}
\begin{lstlisting}[
    language=Python,
    breaklines=true,
    basicstyle=\ttfamily\small,
    keywordstyle=\color{blue},
    commentstyle=\color{green!60!black},
    stringstyle=\color{purple},
    showstringspaces=false,
    numbers=left,
    numberstyle=\tiny\color{gray},
    frame=none
]
    ......
    # Dodge-specific reward
    action = action_state_changes[action_idx]
    if action["action_name"] == "Dodge":
        if player_health_change == 0:
            reward += 2  
        else:
            reward -= 0.5
    ......
    # Combo rewards/penalties
    def calculate_combo_reward(action_history):
        ......
        # Penalty for excessive dodging
        if action_history.count(6) > 15:
            combo_reward -= 5 
        ......
        return combo_reward

    reward += calculate_combo_reward(action_history)

    return reward
\end{lstlisting}

\textbf{Code after optimization}
\begin{lstlisting}[
    language=Python,
    breaklines=true,
    basicstyle=\ttfamily\small,
    keywordstyle=\color{blue},
    commentstyle=\color{green!60!black},
    stringstyle=\color{purple},
    showstringspaces=false,
    numbers=left,
    numberstyle=\tiny\color{gray},
    frame=none
]
    ......
    # Dodge-specific reward
    action = action_state_changes[action_idx]
    if action["action_name"] == "Dodge":
        if player_health_change == 0:
            reward += 0.5  
        else:
            reward -= 0.1
    ......
    # Combo rewards/penalties
    def calculate_combo_reward(action_history):
        ......
        # Penalty for excessive dodging
        if action_history.count(6) > 10:
            combo_reward -= 5 
        ......
        return combo_reward

    reward += calculate_combo_reward(action_history)

    return reward
\end{lstlisting}

We also found that VLM does not always optimize the reward logic. There is also a low probability of misunderstanding, such as making a mistake in the calculation logic of boss health during the optimization process, as shown in the following example:
\begin{lstlisting}[
    language=Python,
    breaklines=true,
    basicstyle=\ttfamily\small,
    keywordstyle=\color{blue},
    commentstyle=\color{green!60!black},
    stringstyle=\color{purple},
    showstringspaces=false,
    numbers=left,
    numberstyle=\tiny\color{gray},
    frame=none
]
    ......
    # Boss health change reward
    boss_health_reduction = 1-prev_state["boss_percentage"]
    if boss_health_change > 0.02:
        reward +=
    ......
\end{lstlisting}

\section{Prompts We Used}
\textbf{Game Environment Analysis}
\begin{lstlisting}[
    language=Python,
    breaklines=true,
    basicstyle=\ttfamily\small,
    keywordstyle=\color{blue},
    commentstyle=\color{green!60!black},
    stringstyle=\color{purple},
    showstringspaces=false,
    numbers=left,
    numberstyle=\tiny\color{gray},
    frame=none
]
env_sys_prompt='''
You are a specialized game environment analyzer with expertise in processing and interpreting video game screenshots. 
Your core capabilities include:
1. Precise scene classification between UI and gameplay environments
2. Detailed visual element extraction and spatial relationship analysis
3. Gameplay situation assessment

Your analysis must be accurate, concise, and focus on actionable information that would be relevant for game AI decision-making.'''

def generate_prompt(game_info):
    prompt = f"""
You are a game AI assistant responsible for analyzing in-game screenshots. Your task is to identify the type of the current screenshot and summarize the key information within it.

There are two types of screenshots:
1. **UI Screen**: Refers to screenshots displaying menus or user interfaces.
2. **Gameplay Screen**: Refers to actual gameplay screenshots, showing characters, enemies, items, and other scene elements.

You need to follow these steps:
1. Determine the screenshot type: Is it a "UI Screen" or a "Gameplay Screen"?
2. If it's a **UI Screen**,
    - extract and summarize the text from the UI, such as options, buttons, etc.
3. If it's a **Gameplay Screen**
    -  First assess the Camera View state: Check if view is too high (excessive sky/trees visible); too low (excessive ground visible); left/right (incomplete road visibility) and road features are clearly visible
    - extract the key information based on the following elements: {game_info.get('Frame_attention')}
    - For enemy detection, use EXACTLY one of these formats:
        * If enemies present: "Enemy detected: [number] enemies at [position]"
        * If no enemies: "No enemy detected"
    - summarize the environment or Point out potential dangers or opportunities
    
Output a your result in the following format:
  screen type is: "<UI Screen or Gameplay Screen>",
  observation is: "<Summary of the content>"

Example output for Gameplay Screen:
screen type is: "Gameplay Screen",
observation is: "
    1. camera view state is: (1) View angle slightly too high - excess sky visible; (2) Road visibility partially blocked on right side
    2. Path details is: Main path heading north through forest
    3. Enemy detected: 2 enemies at front
    4. environment summarize is: Forest path blocked by two enemies with dense vegetation on both sides"

"
"""
    return prompt
\end{lstlisting}

\textbf{Historical Task Summary}
\begin{lstlisting}[
    language=Python,
    breaklines=true,
    basicstyle=\ttfamily\small,
    keywordstyle=\color{blue},
    commentstyle=\color{green!60!black},
    stringstyle=\color{purple},
    showstringspaces=false,
    numbers=left,
    numberstyle=\tiny\color{gray},
    frame=none
]
history_summary_sys_prompt = '''
You are an expert game historian. Your role is to synthesize gameplay history into a concise, informative narrative paragraph that captures key events, strategies, and insights relevant for future decision-making.
'''
def history_summary_prompt(history_logs):
    base_prompt = f"""
Based on the following game history logs, generate a single coherent paragraph (approximately 150 words) that:
- Summarizes the key events chronologically
- Highlights critical decisions and their outcomes
- Identifies important patterns or strategies
- Notes any significant environmental changes
- Includes relevant insights for future tasks

Game History Logs:
{history_logs}

Your summary should be clear, concise, and focused on information that will be most valuable for future task reasoning.
"""
    return base_prompt

\end{lstlisting}

\textbf{Previous Task Reflection}
\begin{lstlisting}[
    language=Python,
    breaklines=true,
    basicstyle=\ttfamily\small,
    keywordstyle=\color{blue},
    commentstyle=\color{green!60!black},
    stringstyle=\color{purple},
    showstringspaces=false,
    numbers=left,
    numberstyle=\tiny\color{gray},
    frame=none
]
task_sys_prompt='''
'You are an expert game analyst specializing in task reflection and evaluation. Your role is to:
1. Analyze all gameplay screenshots and state changes to understand what happened during task execution
2. Evaluate task completion status with concrete evidence
3. Identify and analyze issues at task design, action planning, and execution levels
4. Provide specific recommendations when needed

Always provide detailed, objective analysis following the exact format requested in the prompt.'''

def generate_task_level_prompt(pass_task_info, pass_env_info, current_env_info, pass_action_code):
    base_prompt = f"""Analyze the previous task execution using the following information:

    1. Task Information:
    {pass_task_info}

    2. Environment States:
    - Before task execution: {pass_env_info}
    - After task execution: {current_env_info}

    3. Action Design:
    - Planned action list and Execution code: 
    {pass_action_code}

    Please conduct your analysis in these sequential steps and provide a detailed response in the following format:

    1. VISUAL ANALYSIS
    Provide a clear description of:
    - What happened during the task execution based on all the gameplay screenshots
    - Key UI changes (if in UI screens), character movements, interactions observed, and Notable changes in environment states
    {" - Changes between initial and final maps (The last two pictures)" if has_map else ""} 

    2. TASK COMPLETION EVALUATION
    State clearly:
    - Whether the task was successfully completed
    - Specific evidence from screenshots or state changes supporting your conclusion

    3. ISSUE ANALYSIS (if any problems occurred)
    Analyze at three levels:
    a) Task Design Level
        - Any issues with task design given the game state
        - Problems with task objectives or prerequisites

    b) Action Planning Level
        - Issues with the planned action sequence
        - Problems with action strategy or logic

    c) Action Execution Level
        - Problems with specific control inputs
        - **Issues with duration of actions**

    4. NEXT STEP RECOMMENDATION
    If task failed:
    - Specific suggestions to complete the task in the **CURRENT** state

    If task succeeded:
    - Simply state that the task was completed successfully and no modifications are needed

    Please provide your analysis in the following format:
VISUAL ANALYSIS:
<Describe the sequence of events observed in gameplay screenshots, including UI changes (if in UI screens), character actions, and any significant state changes>
{" <Describe any relevant changes observed between initial and final maps>" if has_map else ""}

TASK COMPLETION EVALUATION:
Status: <SUCCESS/FAILURE>
Evidence: <List specific observations from screenshots or state changes that support your status determination>

ISSUE ANALYSIS:
Task Design Level:
<Evaluate if there are any issues with how the task was designed and specified. If no issues, explicitly state that>

Action Planning Level:
<Analyze if the planned sequence of actions was appropriate and complete. Identify any logical gaps or problems>

Action Execution Level:
<Assess if there were any issues with the specific implementation of actions, such as timing or input problems>

NEXT STEP RECOMMENDATION:
<If task failed: Provide specific suggestions for task completion given the current state>
<If task succeeded: Simply state that the task was completed successfully and no modifications are needed>

"""

    return base_prompt
\end{lstlisting}

\textbf{Action Design Reflection}
\begin{lstlisting}[
    language=Python,
    breaklines=true,
    basicstyle=\ttfamily\small,
    keywordstyle=\color{blue},
    commentstyle=\color{green!60!black},
    stringstyle=\color{purple},
    showstringspaces=false,
    numbers=left,
    numberstyle=\tiny\color{gray},
    frame=none
]
action_sys_prompt='''
You are an expert game action analyst specializing in analyzing and improving game control implementations. Your role is to:
1. Analyze gameplay screenshots to understand the execution effects of each action
2. Evaluate action code design and implementation quality
3. Provide reusable insights for similar actions in the future
4. Suggest specific improvements for action code design

Always provide detailed, objective analysis following the exact format requested in the prompt.
'''

def generate_action_level_prompt(pass_task_info, pass_action_code):
    base_prompt = f"""Analyze the previous action execution using the following information:

    1. Screenshot Sequence Rules:
    - For WASD movement actions lasting over 2 seconds:
        * Screenshots are captured every 2 seconds during the movement
    - For all other key/mouse actions:
        * Only two screenshots are captured: one before and one after the action
    This helps track continuous movements and precise action effects.

    2. Task Context:
    {pass_task_info}

    3. Action plan and code list:
    {pass_action_code}

    Please conduct your analysis in these sequential steps and provide a detailed response in the following format:

    1. ACTION EXECUTION ANALYSIS
    For each action in the sequence, analyze:
    - Initial state and final state from screenshots
    - Whether the action achieved its intended effect
    - Timing and smoothness of execution
    - Any unexpected behaviors or side effects

    2. ACTION CODE EVALUATION
    For each action implementation, evaluate:
    - Appropriateness of key/mouse mapping choices
    - Timing duration settings
    - Action sequence coordination
    - Code efficiency and reliability

    3. SUCCESS/FAILURE ANALYSIS
    For each action, determine:
    - Whether it succeeded or failed
    - Root causes of any failures:
        a) Input mapping issues
        b) Timing problems
        c) Sequence coordination issues
        d) Environmental factors

    4. REUSABILITY ANALYSIS
    Analyze each action's potential for reuse:
    - Common scenarios where this action pattern could apply
    - Required prerequisites and conditions
    - Potential adaptations needed for different contexts
    - Limitations and constraints

    5. IMPROVEMENT RECOMMENDATIONS
    Provide specific suggestions for:
    - Better key/mouse mapping choices
    - Optimal timing parameters
    - Enhanced sequence coordination
    - More robust implementation patterns
    
    Note that:
1. output will be directly evaluated using Python eval(), so it must be a valid Python list of dicts
2. No additional text or explanation should be added between or after these sections
    After completing your analysis, output a list of dictionaries in the following format:

    ```python
    [
        {{
            "action_name_description": "<original action description from action_name_description>",
            "action_code": "<corresponding action code tuple from action_code>",
            "reflection": {{
                "execution_analysis": "<summary of execution analysis>",
                "code_evaluation": {{
                    "status": "<SUCCESS/PARTIAL SUCCESS/FAILURE>",
                    "quality_analysis": "<implementation quality summary>"
                }},
                "success_failure_analysis": "<detailed analysis of what worked/didn't work>",
                "reusability": {{
                    "applicable_scenarios": "<list of potential reuse cases>",
                    "prerequisites": "<required conditions>",
                    "limitations": "<known constraints>"
                }},
                "improvements": "<specific suggestions for implementation improvements>"
            }}
        }},
        # ... repeat for each action
    ]
    ```

    Ensure your response ends with this structured list for easy parsing. Format it exactly as shown above.
    """
    return base_prompt

\end{lstlisting}

\textbf{Task Planning}
\begin{lstlisting}[
    language=Python,
    breaklines=true,
    basicstyle=\ttfamily\small,
    keywordstyle=\color{blue},
    commentstyle=\color{green!60!black},
    stringstyle=\color{purple},
    showstringspaces=false,
    numbers=left,
    numberstyle=\tiny\color{gray},
    frame=none
]
task_planner_sys='''You are an intelligent game AI assistant specializing in strategic task planning and execution. 

Key Responsibilities:
1. Analyze game situations comprehensively considering:
   - Current state and environment
   - Historical context and past experiences
   - Game objectives and constraints

2. For ALL tasks (not just movement), provide:
   - Clear, specific, and actionable objectives
   - Precise success criteria
   - Required resources or conditions
   - Risk assessment and mitigation strategies

3. For movement-related tasks, MUST provide precise location descriptions using:
   - Relative position to character (using character height as scale)
   - Directional instructions (up/down/left/right or compass directions)
   - Safe path recommendations considering terrain

4. Special Considerations:
   - Prioritize agent safety and objective completion
   - Balance exploration with risk management
   - Adapt strategy based on previous task outcomes
   - Consider resource management and efficiency

Your task is to make informed decisions that progress game objectives while maintaining agent safety and efficiency.
'''

def construct_task_prompt(current_frame, pass_task_history_summary, pass_task_reflection, env_info, game_info, step):

    base_prompt = f"""
Analyze the current situation and plan the most appropriate next task considering:

1. Game Objectives: 
{game_info.get('Global_task')}
2. Additional Task Context: {game_info.get('additional_task_info4_task_plan')}

Your design task should be broken down into the following specific Available Controls:
{game_info.get('control_info')} 

Required Analysis Steps:
1. Evaluate current environment and state
2. Consider historical context and lessons learned
3. Assess risks and opportunities
4. Determine priority actions"""


    current_state = f"""
Current Environment Status:
{env_info}"""

    if step == 1:
        analysis_prompt = f"""{base_prompt}
{current_state}

This is the initial step. Focus on understanding the current situation and establishing a safe starting point."""

    else:
        history_context = f"""
Historical Context:
Task History Summary: {pass_task_history_summary}

Previous Task Reflection: {pass_task_reflection}"""
        analysis_prompt = f"""{base_prompt}
{current_state}
{history_context}

Consider:
1. Previous task outcomes and lessons learned
2. Current environmental constraints
3. Progress toward game objectives
4. Safety and risk management"""

    output_format = """
Based on your analysis, provide your response in the following format:

reasoning process:
    1. Current State Analysis: "<analyze current environment and immediate situation>"
    2. Historical Context: "<analyze relevant history and reflections>"
    3. Strategic Evaluation: "<evaluate opportunities, risks, and priorities>"

task details:
    goal: "<specific, actionable objective>"

    location details:
        - screen_position: "<describe target position. Example: '3 meters to the right'>"
    key_requirements: "<essential conditions or resources needed>"
    success_criteria: "<main condition that must be met>"


Note: 
- The direction of camera adjustment **MUST** be consistent, and there should be no angle that switches left and then right, or up and then down
- For movement-related tasks, always specify both screen-relative positions (using character height as scale). For non-movement tasks, mark position fields as 'N/A' if not relevant."""

    return analysis_prompt + output_format
\end{lstlisting}
\textbf{Action Planning}
\begin{lstlisting}[
    language=Python,
    breaklines=true,
    basicstyle=\ttfamily\small,
    keywordstyle=\color{blue},
    commentstyle=\color{green!60!black},
    stringstyle=\color{purple},
    showstringspaces=false,
    numbers=left,
    numberstyle=\tiny\color{gray},
    frame=none
]
action_prompt =  f"""Based on the task you just planned, break it down into specific executable actions.
Please list the specific actions needed to complete this task.

Available Controls:
{game_info.get('control_info')}  

Note that:
1. output will be directly evaluated using Python eval(), so it must be a valid Python list
2. No additional text or explanation should be added between or after these sections
3. Ignore actions such as' wait 'and' observe' that cannot be associated with available controls
4. Action list is *no longer than 5!!*.

Output Format MUST be exactly as follows:
["Action1: <action name> - <detailed description including precise measurements and requirements>","Action2: <action name> - <detailed description including precise measurements and requirements>", ...]

"""
\end{lstlisting}
\textbf{Action Construction}
\begin{lstlisting}[
    language=Python,
    breaklines=true,
    basicstyle=\ttfamily\small,
    keywordstyle=\color{blue},
    commentstyle=\color{green!60!black},
    stringstyle=\color{purple},
    showstringspaces=false,
    numbers=left,
    numberstyle=\tiny\color{gray},
    frame=none
]
action_sys_prompt = '''
You are an expert game AI action planner specializing in converting high-level tasks into precise, executable action sequences.

Key Responsibilities:
1. Convert task descriptions into specific control sequences
2. Ensure accurate timing and duration for each action
3. Maintain action safety and efficiency
4. Generate properly formatted action code that can be directly executed

Important Guidelines:
1. All outputs must be in valid Python dictionary list format
2. Each action must include both description and corresponding control code
3. Control codes must use only valid game controls
4. All durations must be reasonable and safe
'''

def generate_action_prompt(game_info,reason_task,action_plan):
    prompt = f"""
You are an expert game AI action planner specializing in converting high-level action into precise, executable action sequences.

Your current task:
{reason_task}

The action plan for task:
{action_plan}

Available Controls:
{game_info.get('control_info')}  

Additional Action Information:
{game_info.get('additional_action_info')} 

Requirements:
1. Convert each action into specific control sequences
2. Provide both action description and control code
3. Ensure precise timing for each control input
4. Consider safety in all actions

Output Format MUST be exactly as follows:
[
    {{
        "action_name_description": "<original action description>",
        "action_code": [("<key>", <duration>), ...]
    }},
    ...
]

Example Output:
[
    {{
        "action_name_description": "Move Forward - Move 3 meters forward",
        "action_code": [("W", 3.0)]
    }},
    {{
        "action_name_description": "Jump and Interact - Jump over obstacle and press button",
        "action_code": [("SPACE", 0.1), ("E", 0.1)]
    }}
]

Note:
1. Output will be evaluated using Python ast.literal_eval()
2. Use only valid control keys: {list(game_info.get('Mapping_info', {}).keys())}
3. All durations must be positive numbers
4. Maintain exact format with no additional text
"""
    return prompt
\end{lstlisting}

\end{document}